\pdfoutput=1

\documentclass[11pt]{article}

\usepackage{ACL2023}
\usepackage{times}
\usepackage{latexsym}
\usepackage{spverbatim}
\usepackage{graphicx}
\usepackage{dirtree}
\usepackage{url}

\usepackage[T1]{fontenc}

\usepackage[utf8]{inputenc}

\usepackage{microtype}

\usepackage{inconsolata}

\title{Creating a Fine Grained Entity Type Taxonomy Using LLMs}

\author{Michael Gunn \\
  \texttt{mwgunn2@illinois.edu} \\\And
  Dohyun Park \\
  \texttt{dohyunp2@illinois.edu} \\\And
  Nidhish Kamath \\
  \texttt{nkamath5@illinois.edu} \\}

\begin{document}
\maketitle
\begin{abstract}
In this study, we investigate the potential of GPT-4 and its advanced iteration, GPT-4 Turbo, in autonomously developing a detailed entity type taxonomy. Our objective is to construct a comprehensive taxonomy, starting from a broad classification of entity types - including objects, time, locations, organizations, events, actions, and subjects - similar to existing manually curated taxonomies. This classification is then progressively refined through iterative prompting techniques, leveraging GPT-4's internal knowledge base. The result is an extensive taxonomy comprising over 5000 nuanced entity types, which demonstrates remarkable quality upon subjective evaluation.

We employed a straightforward yet effective prompting strategy, enabling the taxonomy to be dynamically expanded. The practical applications of this detailed taxonomy are diverse and significant. It facilitates the creation of new, more intricate branches through pattern-based combinations and notably enhances information extraction tasks, such as relation extraction and event argument extraction. Our methodology not only introduces an innovative approach to taxonomy creation but also opens new avenues for applying such taxonomies in various computational linguistics and AI-related fields.
\end{abstract}

\let\thefootnote\relax\footnotetext{All code and taxonomies created can be found at \href{https://github.com/Psynaps/UltraFine-Ontology}{github}}

\section{Introduction}
Recent breakthroughs in large language models, notably GPT-4 \cite{openai2023gpt4}, have ushered in an era of advanced natural language processing tools. These models exhibit extensive general language comprehension capabilities, resulting from their extensive training process. GPT-4, in particular, stands out for its deep understanding of language nuances. Utilizing their language understanding, these large language models have greatly improved the performance of various natural language tasks \cite{mao2023gpteval}. However, despite these advances, these models can struggle to maintain consistency in structured knowledge representation.

Addressing this gap motivates the development of a robust type taxonomy. Historically, type taxonomies, whether small-scale or human-crafted, have been pivotal in various domains, including entity typing, relation extraction, and event argument extraction. However, they often fall into two categories: either limited in scope or manually created, posing a significant challenge in being adapted for wide use. This project explores the possibility of leveraging GPT-4's nuanced language understanding to construct a detailed, fine-grained entity type taxonomy.

Our project begins with an initial exploration, inspired by the granular entity types identified in Ultrafine \cite{choi-etal-2018-ultra}. Next, we detail our comprehensive process of taxonomy development as well as the resulting taxonomy. We delve into the methodology, highlighting the iterative prompting techniques employed to refine and expand the taxonomy. Furthermore, we discuss the potential applications of this fine grained taxonomy, exploring its capability in augmenting existing methods in entity typing, relation extraction, and event argument extraction. This exploration not only showcases the capabilities of GPT-4 in structured knowledge organization but also illuminates new avenues for its application in computational linguistics and AI research.

\section{Background}

The evolution of Large Language Models (LLMs) such as GPT-4 has significantly impacted the field of natural language processing, particularly in the creation and utilization of type taxonomies. These taxonomies play a crucial role in structuring knowledge for complex NLP tasks. Among the pioneering works in this domain is the UltraFine Entity Typing Task\cite{choi-etal-2018-ultra}. This endeavor focuses on discerning fine-grained entity types, advancing the representation of structured knowledge in numerous NLP tasks like relation extraction and event argument extraction. Yet, despite its innovativeness, this method encountered scalability and consistency challenges, particularly in managing a wide spectrum of entity types.

Another significant contribution in this field is the OntoType model\cite{komarlu2023ontotype}. OntoType introduces an ontology-guided, zero-shot fine-grained entity typing system with weak supervision from pre-trained language models. This method represents a substantial advancement in understanding and classifying fine-grained entity types, particularly in how it leverages the pre-existing knowledge embedded in language models.

Our work extends these efforts by exploring the potential of GPT-4 in autonomously creating a fine-grained entity type taxonomy. We build upon the foundational concepts established in the UltraFine Entity Typing Task and integrate them with the advanced capabilities of GPT-4 and GPT-4 Turbo. This integration aims to overcome the limitations faced in previous approaches, such as the UltraFine system's reliance on crowdsourced results and the challenges in scaling up to thousands of entity types.

In our methodology, we utilize GPT-4's language understanding to develop a comprehensive taxonomy, starting from a broad classification and progressively refining it through iterative prompting techniques. This process results in an extensive taxonomy comprising over 5000 nuanced entity types. Our approach not only introduces an innovative way of creating taxonomies, but also demonstrates the practical applications of such a detailed taxonomy in enhancing information extraction tasks. For instance, the taxonomy facilitates the creation of more specific event patterns using a hierarchical entity taxonomy, which is particularly beneficial in event argument extraction and relation extraction tasks.

In summary, our project draws on the foundations laid by previous work in type taxonomies, enriching it through the application of GPT-4's capabilities. We propose a methodology that refines the concept of structured knowledge representation in NLP. While it marks progress in the field, we also acknowledge the complexities and ongoing challenges in creating and applying extensive taxonomies. Our work, therefore, should be seen as a step towards enhancing information extraction tasks, offering insights into how advanced language models can contribute to the development of more nuanced and effective taxonomies in computational linguistics.

\section{UltraFine Entity Typing Task}
We motivate our work by highlighting the entity typing task \cite{choi-etal-2018-ultra} and contrasting how humans and LLMs might fare on it. The task is to predict natural-language phrases that describe an entity \textit{e} contained in a given sentence. Consider the following sentence and the entity `it\textquotesingle s' as an example:

\begin{spverbatim}
“For starters, it’s not an ordinary sun but a Cepheid variable - a giant, pulsating star shining with the light of at least a thousand suns.”
\end{spverbatim}

\vspace{10pt}

What are the applicable types for `it\textquotesingle s' in this sentence? Crowdsourced results from UltraFine \cite{choi-etal-2018-ultra} labeled it as `object', `celestial body', `sun', `star'. An LLM such as GPT-4 can easily provide answers like `sun' or `star' directly, but would miss out on stating the other perfectly relevant entity types. This is where we feel that having an ontology of entity types would greatly improve results. A natural fine grained ontology would contain `sun' under:

\vspace{5pt}

\dirtree{%
.1 Object.
.2 \ldots{}.
.3 Celestial body.
.4 Star.
.5 Sun.
}

\vspace{5pt}

Any method that can infer `it' as a type of `sun' could then, by utilizing the structured knowledge of the ontology, infer all parent types as being applicable too.

From this example, we believe the role of a well structured fine grained type taxonomy in creating extensive multi-labeling of entities is clear. Given a well structured taxonomy like the one we create in this work, the task of providing a thorough list of applicable labels for an entity becomes much easier. In particular, existing simple prompting methods can already output accurate results of more specific type labels, which can then be traced using a type taxonomy to find all applicable labels. 

\section{Methodology}
\subsection{Initial Approach}

In our initial approach, we explored the potential of leveraging a GPT-4 to structure a comprehensive ontology from a provided list of types. We use the list of 10,331 fine grained entity types provided by the Ultra Fine Entity typing task \cite{choi-etal-2018-ultra} and GPT-4 as our LLM. The fundamental strategy involved feeding this exhaustive list to a sufficiently advanced LLM, coupled with an aptly crafted prompt. The objective was to harness GPT-4's extensive world knowledge to systematically organize these entity types into a cohesive ontology.

\begin{figure}[h!]
\centering
\includegraphics[width=0.5\textwidth]{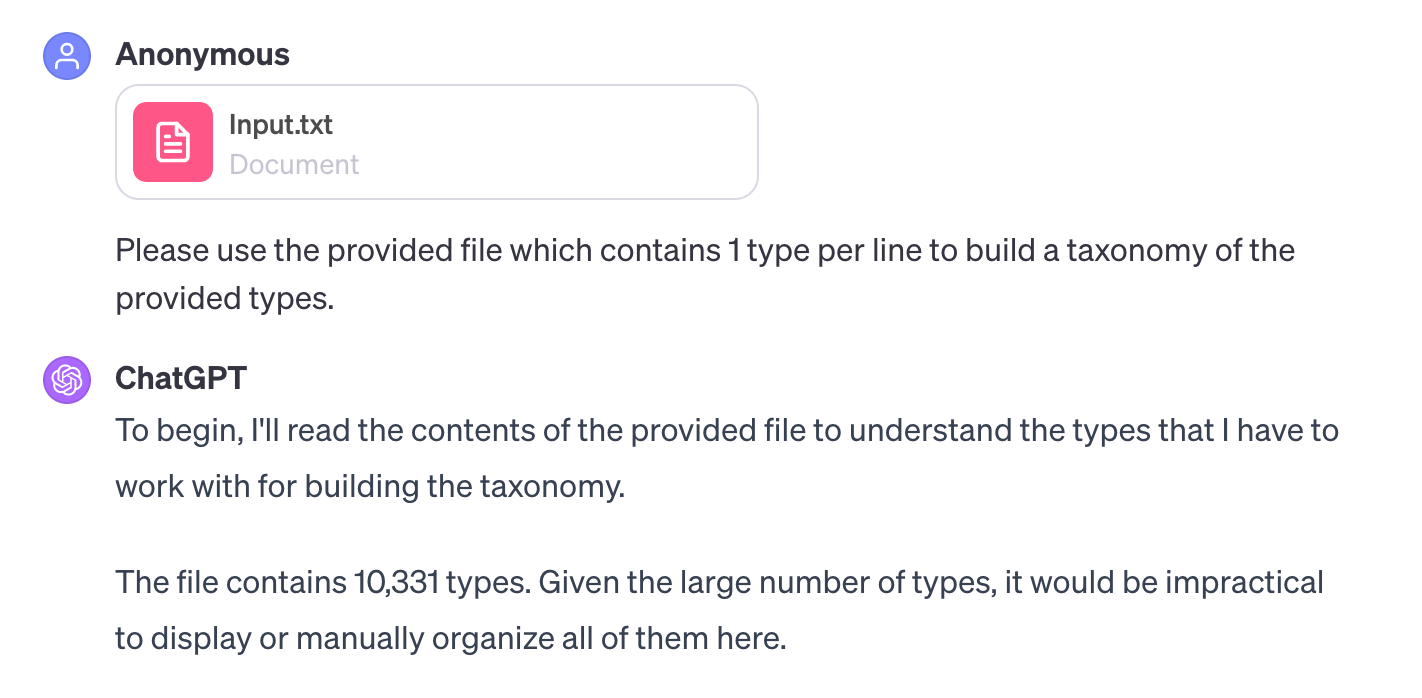}
\caption{Initial Approach Prompt}
\label{fig:init_prompt}
\end{figure}

\subsubsection{Limitations of Initial Approach}\label{init_limitations}

\begin{figure}[!h]
    \centering
    \includegraphics[width=\columnwidth]{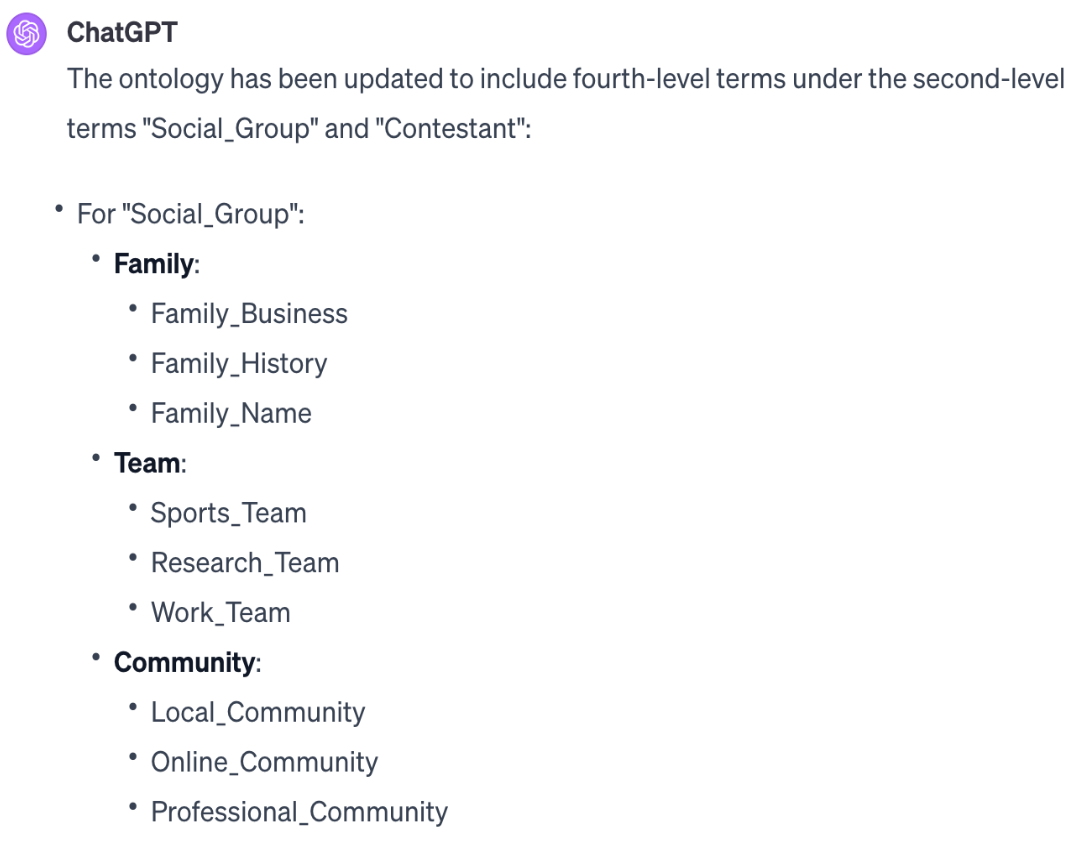}
    \caption{GPT-4 matching phrases on String Similarity}
    \label{fig:string_sim}
\end{figure}

This approach initially showed promise, as GPT-4 was capable of understanding the requirements and generating coherent ontology structures. However, certain limitations hindered further adoption. One significant issue was the occurrence of hallucinations. Despite being prompted to adhere to the provided UltraFine types, GPT-4 sometimes introduced unrelated terms into the ontology. Another challenge involved the method of matching terms based on \textbf{string similarity}. While we aimed to leverage GPT-4's general language understanding to build an ontology, despite our prompts, we found that GPT-4 would tend to group phrases from the UltraFine types if they shared common words, regardless of their semantic context. An example of this was grouping 'Family\_Business', 'Family\_History', and 'Family\_Name' under 'Family' simply due to the shared word. This approach often led to inaccurate grouping as can be seen in figure \ref{fig:string_sim}. Additionally, the unordered nature of the UltraFine types list itself was a limitation. An ideal scenario would involve feeding GPT-4 types that fit into an existing hierarchy. However, the lack of order in the list complicated the construction of a semantically accurate hierarchy, often leading to subpar outputs.

\subsubsection{Initial Solutions}\label{sec:init_sol}

To address the limitations from Section \ref{init_limitations}, we opted for a more structured approach. Given that the Ultrafine types list doesn't provide a clear starting point, we opted to develop a basic, broad level ontology using a selection of key terms. This initial framework serves as a \textit{seed} ontology, which provides a foundational structure for GPT-4 to build upon.

Additionally, we formulated specific prompts aimed at encouraging GPT-4 to generate ontology relations grounded in a deeper comprehension of the meanings behind the terms instead of simple string similarity.

\begin{figure}[!h]
    \centering
    \includegraphics[width=\columnwidth]{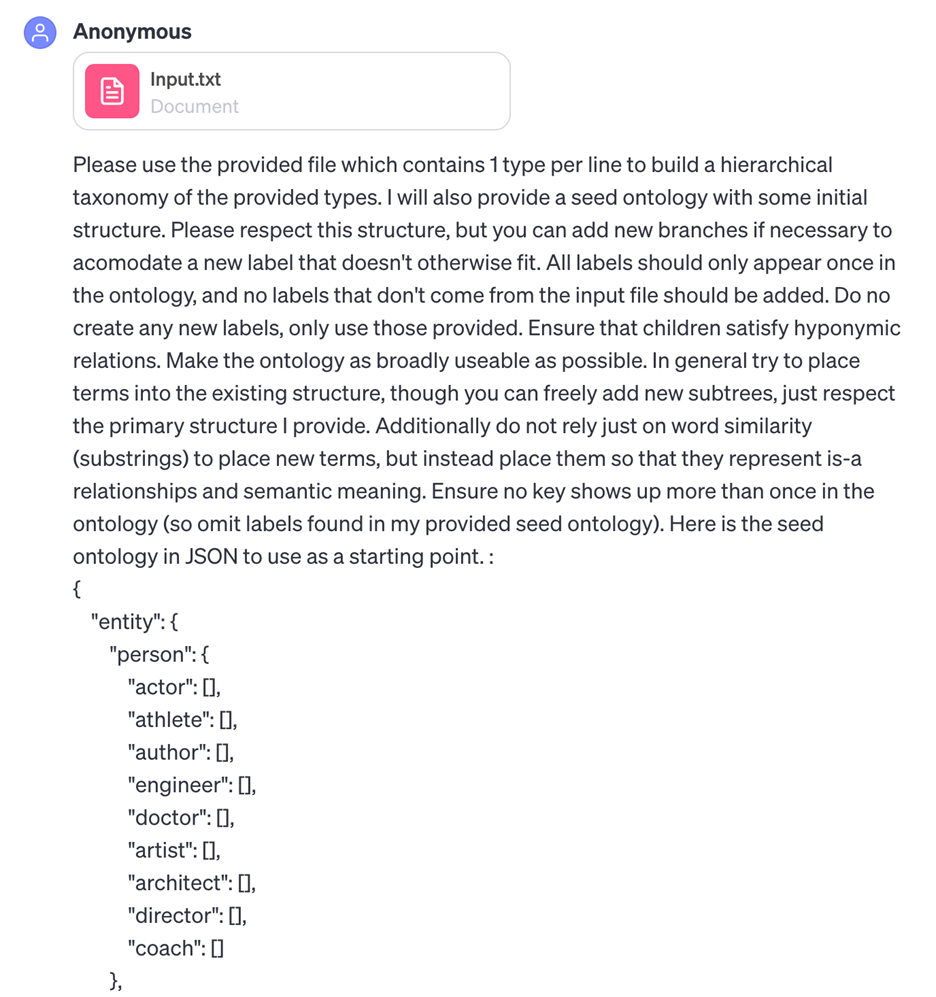}
    \caption{Revised prompt with seed ontology and other explicit instructions}
    \label{fig:seed_ontology}
\end{figure}

Contrary to our expectations, GPT-4 predominantly employed basic string matching rather than demonstrating its general language understanding while building ontologies in this manner. This tendency was particularly evident when the model was tasked with integrating \textit{parent} entities into an existing structure where \textit{child} entities had already been positioned. In such instances, GPT-4 exhibited a tendency to simply append the \textit{parent} entities as a standalone node at the first level, rather than reorganizing the structure for optimal coherence. 

This showed a critical limitation in the model's capacity to dynamically restructure ontologies in response to the sequential presentation of entity types, a constraint notably evident in the context of the unordered UltraFine entity type list. 

These observations suggest that our initial methodology may not be effectively scalable or applicable to extensive and complex entity lists, such as the 10,331 types featured in the UltraFine list or other similar compilations. Consequently, this necessitates the development of a more robust approach for ontology creation with large language models like GPT-4, focusing on customization and scalability.

\subsection{Final Approach}
Despite our initial efforts, as outlined in section \ref{sec:init_sol}, we found the constraints of UltraFine Types more limiting than beneficial for inference. Consequently, we shifted from our initial strategy to a more unrestricted process in our final approach, in which we leveraged the capabilities of GPT-4 without the constraints of the UltraFine types. In this approach, we do not limit types in our ontology to UltraFine types list, and we allow the same type to be able to be inserted in the ontology multiple times as well. We do not provide a Seed Ontology, allowing GPT-4 to structure the ontology as needed. We develop an iterative process to expand an ontology, allowing us to fully utilize GPT-4's language understanding to add new types to our ontology.

\section{Iterative Process}

\begin{figure}[h!]
    \centering
    \includegraphics[width=0.5\textwidth]{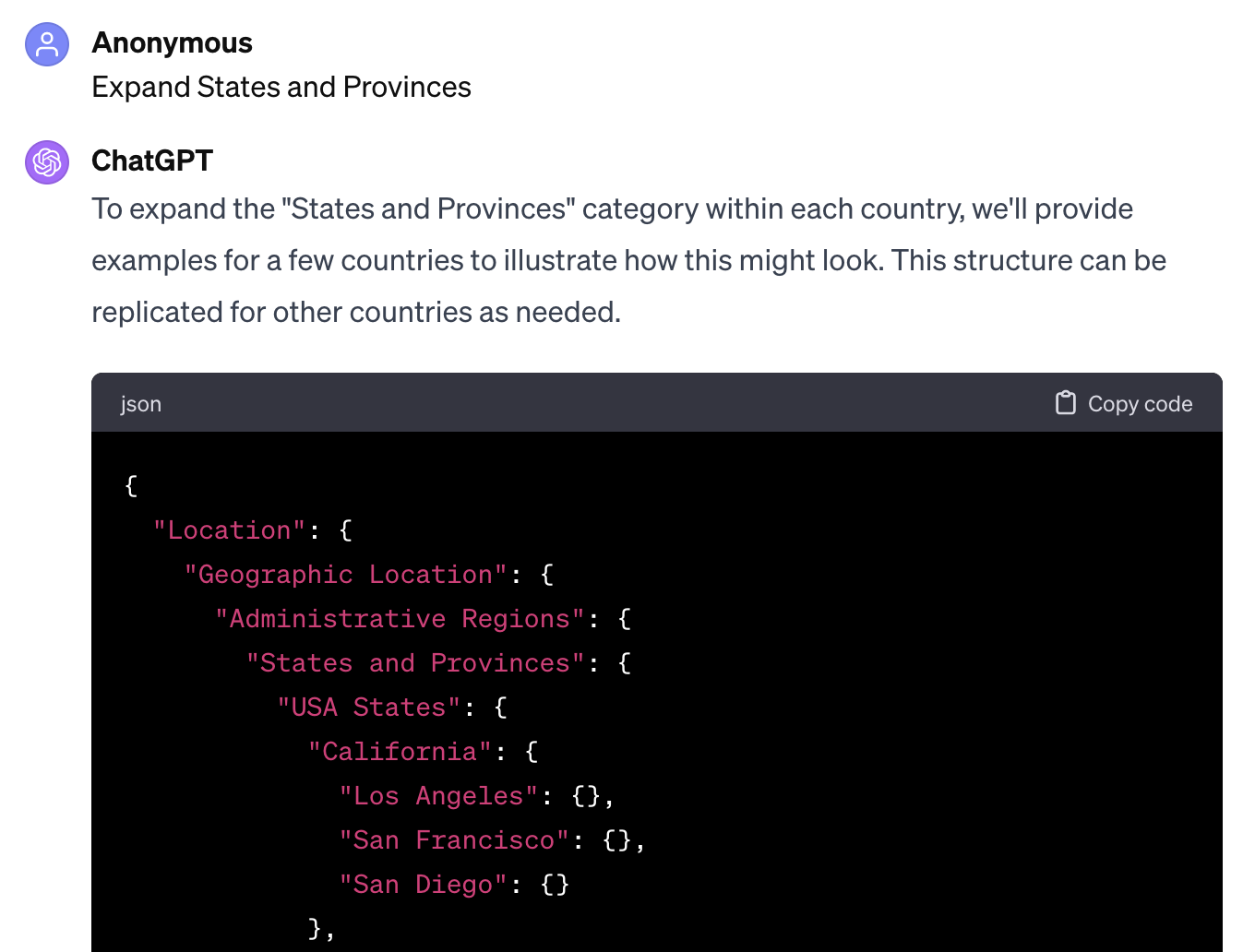}
    \caption{Prompt expanding specific subtree}
    \label{fig:expand-subtree}
\end{figure}

GPT-4's ability to utilize file storage for ontology maintenance simplifies the process of keeping a previous ontology in context. This feature was pivotal in designing an iterative process for building our ontology, enabling us to add new types seamlessly. Each iteration involved presenting GPT-4 with an existing ontology, either from a prior generation or uploaded as a file. With the previous ontology, we now prompt GPT-4 to expand our ontology. We employed two main methods for prompting: one was requesting the expansion of a specific subtree, and the other involved suggesting a new type for GPT-4 to integrate into the most fitting subtree. This approach ensured a dynamic evolution of our ontology, achieving both broad coverage and detailed categorization. Focusing on expanding one subtree (or multiple subtrees of the same type) at a time, as illustrated in figure \ref{fig:expand-subtree}, resulted in faster, more accurate expansions. Another significant advantage of our approach is the ability for multiple team members to concurrently work on different subtrees, enhancing the efficiency of the ontology expansion. This allows each person to focus on a specific area, diversifying the development process and contributing unique perspectives to various parts of the ontology. Additionally, the flexibility to add any desired subtree into the ontology furthers its comprehensiveness, ensuring that our ontology is not only expansive but also tailored to our specific project needs.

\subsection{Extending to Domain Specific Taxonomies}

While the iterative process of building ontologies leverages GPT-4's general language understanding capabilities, it can be further enhanced to incorporate domain-specific knowledge when constructing relevant branches. A recent feature of GPT-4, the \textit{GPT Builder}, enables users to directly feed domain-specific knowledge sources into GPT-4. While untested in this work, utilizing this feature could allow for the creation of more accurate and relevant ontologies within specific fields. Even without a GPT-4-specific feature, our iterative process permits the expansion of specific branches through prompts that can be enhanced with additional knowledge. For example, in the medical context, medical research papers and journals can be input either through the GPT-4 knowledge source or within each prompt. This tailored approach, where domain knowledge is provided with each prompt, can lead to the creation of high-quality and accurate domain-specific ontologies.

\section{Evaluation}
\begin{figure}[t!]
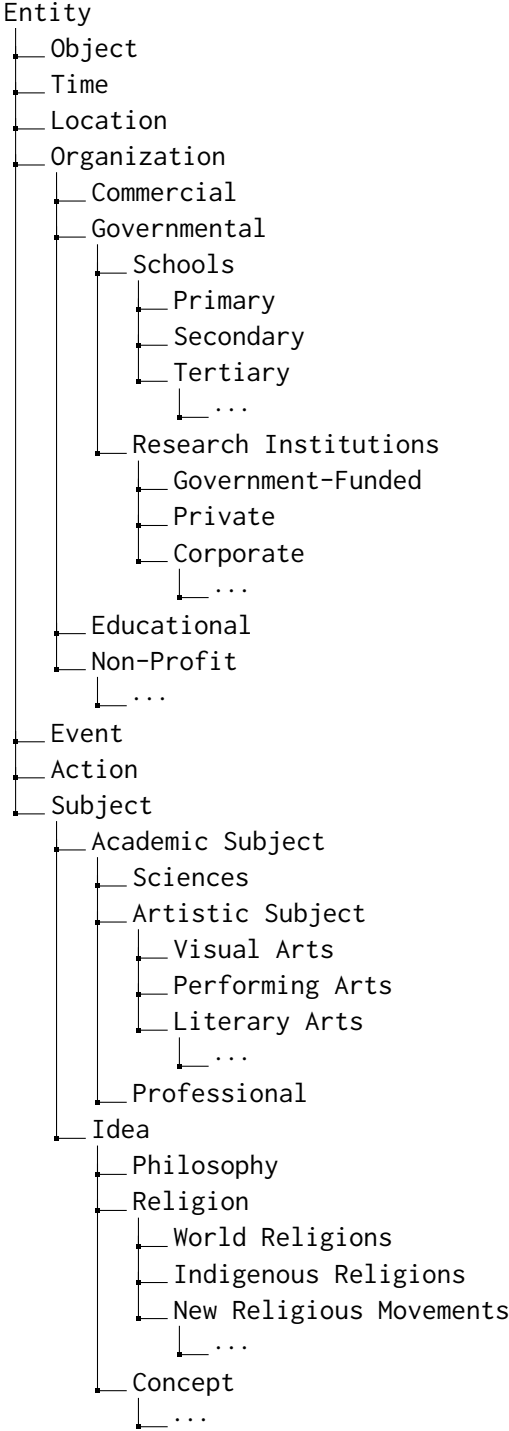

    \dirtree{%
    .1 Entity.
    .2 Object.
    .2 Time.
    .2 Location.
    .2 Organization.
    .3 Commercial.
    .3 Governmental.
    .4 Schools.
    .5 Primary.
    .5 Secondary.
    .5 Tertiary.
    .6 $\cdots$.
    .4 Research Institutions.
    .5 Government-Funded.
    .5 Private.
    .5 Corporate.
    .6 $\cdots$.
    .3 Educational.
    .3 Non-Profit.
    .4 $\cdots$.
    .2 Event.
    .2 Action.
    .2 Subject.
    .3 Academic Subject.
    .4 Sciences.
    .4 Artistic Subject.
    .5 Visual Arts.
    .5 Performing Arts.
    .5 Literary Arts.
    .6 $\cdots$.
    .4 Professional.
    .3 Idea.
    .4 Philosophy.
    .4 Religion.
    .5 World Religions.
    .5 Indigenous Religions.
    .5 New Religious Movements.
    .6 $\cdots$.
    .4 Concept.
    .5 $\cdots$.
    }
    \caption{GPT-4 Generated Ontology subtrees}
    \label{fig:ontology}
\end{figure}

Utilizing the iterative process, we were able to generate an ontology with over 5000 types, reaching depths up to 10. This method proved to be highly scalable, allowing for the addition of any necessary subtree branches. As illustrated in figure \ref{fig:ontology}, our system efficiently categorizes a diverse range of types. Particularly noteworthy is the 'organization' subtree, showcasing our method's ability to delve into detailed categorizations. This illustrates the model's strength in grasping general language concepts and generating intricate, fine-grained ontological structures.

In our proof of concept, we did not expand all possible branches to as deep as possible, but the technique we employed has no limitation for scalability that we observed. GPT-4 is able to read JSON files directly and reference their contents for later expansions. This means that the approach we employed in principle should extend into the many tens of thousands of entity types with no problems. Additionally, we believe the recursive expansion technique we utilized even allows for branch expansion to be performed past the entire taxonomy being able to fit inside the 128k context window of GPT-4 Turbo. It would be possible to pass only the relevant subset of the entire taxonomy into the model in order to allow branch expansion to be performed essentially without limit. 

Overall the quality of the resulting taxonomy that we observed from our approach subjectively was extremely high. All of the hierarchical divisions created made logical sense and were accurate to real world relationships between entities. We were not able to manually check every single entity type in our resulting taxonomy, but all that we did examine were of high quality. 

\subsection{Limitations}
The iterative prompting approach that we developed and employed to create our taxonomy using GPT-4 was very effective, but we did observe and theorize some potential limitations and shortcomings of the method. One such limitation comes from the manner in which we perform branch expansion. We decided to construct the taxonomy using a hierarchical iterative deepening approach, which involved first creating a high level structure of the most general categories for entities, then recursively expanding each branch to obtain more fine grained types and hierarchical structures. In practice this meant that in order to expand a given branch of the taxonomy, we would prompt it to generate a more detailed version of that branch. The resulting output branch then replaced the previous branch entirely. This branch replacement strategy allowed simple expansion of any given branch of the taxonomy, but it means that the selection of which branches should be expanded next is left up to the user. 

Another limitation that we observed in our method comes from the allowance of repeated terms in different places in the taxonomy. From our first approach using the type list from UltraFine \cite{choi-etal-2018-ultra} we observed that GPT-4 could sometimes place the same term in multiple places in the taxonomy. We initially did not feel this was appropriate, but upon further consideration recognized that in a highly extensive type taxonomy, certain terms should occur in multiple sections as they have multiple uses across different areas. One example of this is the term "technology". In an ideal extensive taxonomy the term "technology" should likely occur under the categories of professional occupations, as a subset of objects, under infrastructure, and likely multiple other branches as well. As a result, in our final approach in which we constructed the taxonomy from scratch without restriction on the terms to be used, we relaxed the constraint on term uniqueness.

In general we feel that this decision did allow for a more useful and informative taxonomy, but it does introduce a complication for later applications of the taxonomy. The multiple occurrences of the same term in the taxonomy means that performing a task like entity typing using our taxonomy would necessitate a more detailed description of each type than just its individual label. This problem can be addressed in a number of ways, but the most obvious is to make the resulting type label not just be the final type label from the taxonomy, but instead the entire branch of the taxonomy that ends in that type. Thus instead of "technology" as a type label, one would obtain "Entity / Object / Physical Object / Non-living Beings / Artificial objects / Infrastructure / Technology". The handling of duplicate terms in taxonomy applications does introduce some challenges, but we believe that it is worth those costs in order to have a more informative and unrestricted type tree. 

\section{Ontology Combinations}
\begin{figure}
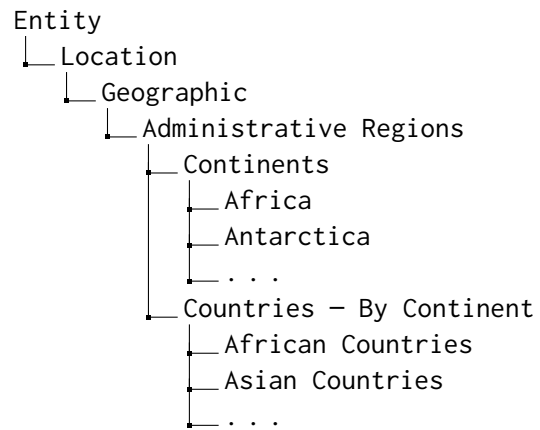

    \dirtree{%
    .1 Entity.
    .2 Location.
    .3 Geographic.
    .4 Administrative Regions.
    .5 Continents.
    .6 Africa.
    .6 Antarctica.
    .6 \ldots{}.
    .5 Countries --- By Continent.
    .6 African Countries.
    .6 Asian Countries.
    .6 \ldots{}.
    }
    \caption{Repetition in Ontology}
    \label{fig:repetition}
\end{figure}

The GPT-4 generated ontology, while detailed, suffered from excessive granularity, leading to heavy repetition within similar but different categories. For example, as can be seen in figure \ref{fig:repetition}, the name of each country can be seen being repeated in both \textbf{Countries} and \textbf{Continents}. Furthermore, the name of each country can be seen in totally separate subtrees such as \textbf{mountains}, where categories such as \textbf{american mountains}, \textbf{canadian mountains} can be seen. Such repetition can hinder search within an ontology, and make it overwhelming for users. To overcome these deficiencies, we define new ontology combinations, such that new logical categories can be defined without having excessive repetition. 

Following our example from figure \ref{fig:repetition}, we can define a new combination : $\textbf{Continent} + \textbf{Country} \rightarrow \textbf{Country} \ \textit{in} \ \textbf{Continent}$. This leads to expressions like \textbf{'Countries in Europe'} or \textbf{'Asian Nations'}, reducing repetition. Similarly, defining the combination $\textbf{Country} + \textbf{fauna}$ leads to expressions such as \textbf{'Australian Mammals'} or \textbf{'Indonesian Tiger'}.

This approach of defining ontology combinations makes generating ontologies simpler while also making it easier for users to find relevant information and understand the relation between the different categories.

\section{Practical Applications}

There are many tangible benefits of employing a fine-grained ontology in NLP and AI tasks, and given our high quality ontology, there are many practical applications where our ontology can be used. Using our ontology can significantly enhance event argument extraction, enabling the identification of more nuanced event patterns that traditional methods may overlook. For the task of entity typing, we believe that the approach introduced in OntoType\cite{komarlu2023ontotype} is extendable to use a fine grained type taxonomy. Their method of performing flat type classification at each level of the tree can be directly extended to work with a deeper and more extensive taxonomy like the one we created. The only modification we foresee in employing their method with our larger taxonomy is to allow multiple subbranches to be explored concurrently in finding the final type label using something like beam search. This modification allows the approach of OntoType to handle any challenging or opaque decisions, resulting from the broader structure of our taxonomy, which must be made to arrive at the most specific label in our taxonomy for a given entity.

In addition to the task of entity typing, an extensive taxonomy proves invaluable in pattern-based relation extraction and exploring meta-patterns in data. Using our ontology with fine-grained entity types would make relation extraction significantly simpler as type-level patterns describe relations well. For example, the ability to discern complex relations like \textbf{\$President}, \textbf{leader}, \textbf{\$Country} illustrates the depth of understanding a quality ontology can provide.

\section{Future Work}
We believe the methods explored in this work can serve as a foundation and motivation to explore the general approach of utilizing LLMs like GPT-4 to build structured knowledge representations, specifically in the form of extensive type taxonomies. This work has demonstrated that from its training, GPT-4 and GPT-4 Turbo have obtained a deep understanding of natural language and the structured hierarchical relationships between entities. We believe that our iterative prompting approach to taxonomy creation can easily be extended to create fine grained type taxonomies for specific domains. If successful, this process could make it tractable to build structured hierarchical tree representations for topics and entities in both existing and emerging domains in an automated fashion.

Despite the difficulty of manual creation, this task has been seen as important enough to motivate the creation of large specialized domain type taxonomies such as the PubMed taxonomy \cite{pubmedTaxonomy}. Taxonomies like that one allow more informative structuring of academic literature and entities within the field. We believe that iterative prompting to build taxonomies should be explored in the context of the construction of specialized branches of a larger general taxonomy like the one we created, or the construction of entirely new taxonomies for specific fields. This process could involve the same zero shot prompting style we utilized, or potentially utilize providing relevant domain specific documents as grounding for the model to utilize when creating a new specialized taxonomy. Retrieval augmented generation (RAG) \cite{lewis2021retrievalaugmented} has shown success in many other contexts, and the application of relevant context using domain documents has become even easier with the new capability of GPT-4 to ingest existing document formats directly and fit them within its massive context window. 

In addition to creating taxonomies, the area of utilizing fine grained entity taxonomies is also ripe for further exploration. Extensive taxonomies such as the one created in this work can be used in existing ontology guided approaches \cite{komarlu2023ontotype} \cite{qu2017weaklysupervised} to increase granularity, as well as applied to new tasks in information extraction that have yet to employ structured taxonomies.  

This work surely only scratches the surface of exploring the capability of LLMs such as GPT-4 in creating and utilizing structured knowledge. We feel the technique we developed and the high quality output we were able to produce using it can be extended and utilized in future work.

\section{Conclusion}
In this work we aimed to explore the potential of using GPT-4 to create a fine grained type taxonomy. We believe that structured knowledge representation and consistency remains a shortcoming of existing LLMs, and extensive taxonomies are one way to address that. We demonstrated that GPT-4 has obtained significant understanding of the structural and hierarchical relations between real world entities and concepts. Utilizing the cutting edge file input and structured output capability of GPT-4, we were able to construct an extensive type taxonomy using iterative prompting that is broadly applicable, high quality, and highly extensible. In the process we also developed an approach that can be applied to construct or even augment existing domain specific type trees. Additionally we explored some of the many potential applications for an extensive type taxonomy. These range from extensions of existing methods that already utilize ontologies, to methods of pattern and branch combination approaches to extending taxonomies. Lastly we consider the potential application of structured type trees to existing information extraction tasks. The exploration of the capabilities of massive and highly capable LLMs like GPT-4 and future models is in its early days, and we hope that this work can serve as inspiration for the further utilization of these new powerful tools.

\bibliography{references}
\bibliographystyle{acl_natbib}

\section{Appendix}
\label{sec:appendix}

\subsection{Ontology Depth}
\begin{figure*}
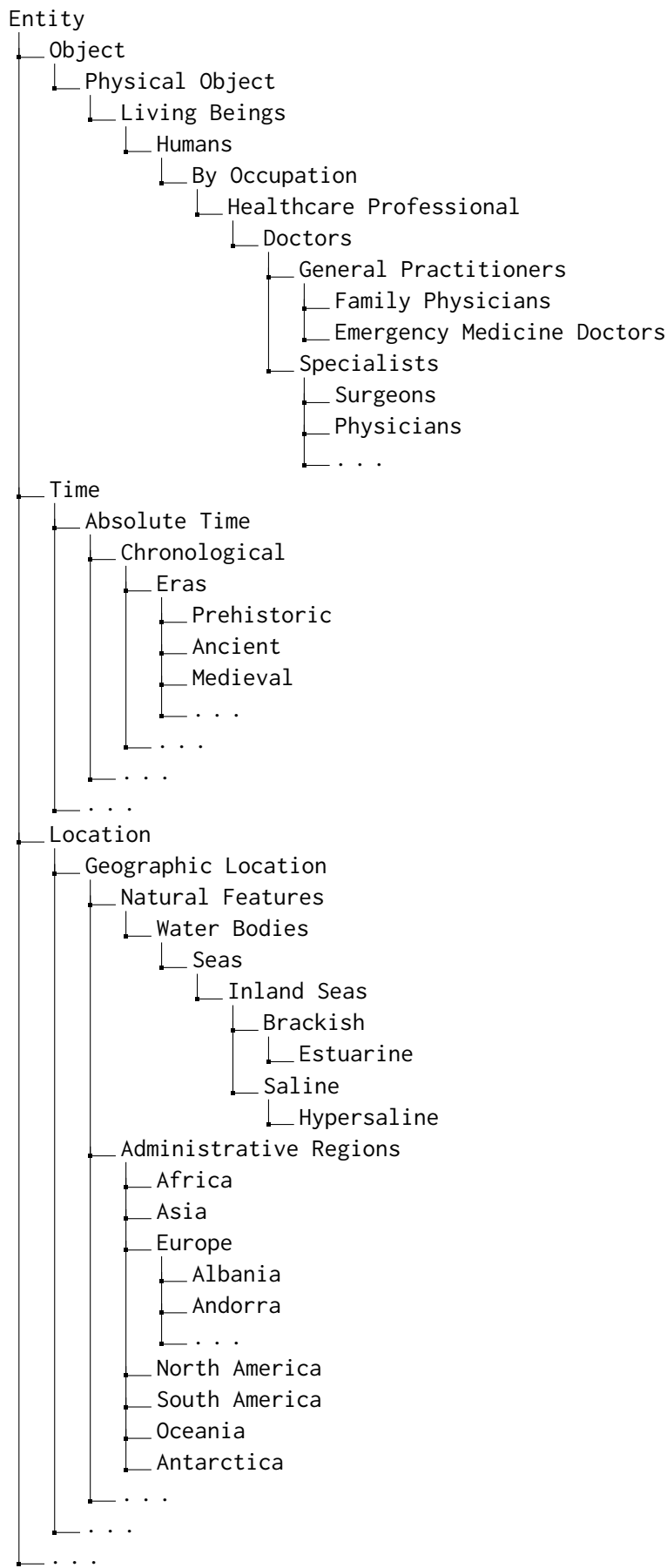

\dirtree{%
.1 Entity.
.2 Object.
.3 Physical Object.
.4 Living Beings.
.5 Humans.
.6 By Occupation.
.7 Healthcare Professional.
.8 Doctors.
.9 General Practitioners.
.10 Family Physicians.
.10 Emergency Medicine Doctors.
.9 Specialists.
.10 Surgeons.
.10 Physicians.
.10 \ldots{}.
.2 Time.
.3 Absolute Time.
.4 Chronological.
.5 Eras.
.6 Prehistoric.
.6 Ancient.
.6 Medieval.
.6 \ldots{}.
.5 \ldots{}.
.4 \ldots{}.
.3 \ldots{}.
.2 Location.
.3 Geographic Location.
.4 Natural Features.
.5 Water Bodies.
.6 Seas.
.7 Inland Seas.
.8 Brackish.
.9 Estuarine.
.8 Saline.
.9 Hypersaline.
.4 Administrative Regions.
.5 Africa.
.5 Asia.
.5 Europe.
.6 Albania.
.6 Andorra.
.6 \ldots{}.
.5 North America.
.5 South America.
.5 Oceania.
.5 Antarctica.
.4 \ldots{}.
.3 \ldots{}.
.2 \ldots{}.
}
    \caption{More examples from our ontology covering more depth and breadth}
    \label{fig:example-ontology-appendix}
\end{figure*}

\end{document}